%% file: main.tex
\documentclass[10pt,twocolumn,letterpaper]{article}

\usepackage{cvpr}              

\usepackage{color}

\definecolor{cvprblue}{rgb}{0.21,0.49,0.74}
\usepackage[pagebackref,breaklinks,colorlinks,citecolor=cvprblue]{hyperref}
\usepackage{gensymb}

\def\confName{CVPR}
\def\confYear{2024}

\usepackage{caption}
\usepackage{subcaption}
\usepackage{booktabs}
\usepackage{graphicx}
\usepackage{amssymb}
\usepackage{booktabs}
\usepackage{multirow}
\usepackage[table]{xcolor}
\usepackage[accsupp]{axessibility}

\author{Lingjun Zhao\textsuperscript{*}, Jingyu Song\textsuperscript{*\dag}, Katherine A. Skinner\\
University of Michigan, Ann Arbor, MI USA\\
\tt\small \{lingjunz,jingyuso,kskin\}@umich.edu
}

\begin{document}
\title{CRKD: Enhanced Camera-Radar Object Detection with Cross-modality Knowledge Distillation}





\date{October 2023}

\maketitle

\begin{abstract}
    In the field of 3D object detection for autonomous driving, LiDAR-Camera (LC) fusion is the top-performing sensor configuration. Still, LiDAR is relatively high cost, which hinders adoption of this technology for consumer automobiles. Alternatively, camera and radar are commonly deployed on vehicles already on the road today, but performance of Camera-Radar (CR) fusion falls behind LC fusion. In this work, we propose Camera-Radar Knowledge Distillation (CRKD) to bridge the performance gap between LC and CR detectors with a novel cross-modality KD framework. We use the Bird's-Eye-View (BEV) representation as the shared feature space to enable effective knowledge distillation. To accommodate the unique cross-modality KD path, we propose four distillation losses to help the student learn crucial features from the teacher model. We present extensive evaluations on the nuScenes dataset to demonstrate the effectiveness of the proposed CRKD framework. The project page for CRKD is \url{https://song-jingyu.github.io/CRKD}.
    \end{abstract}
    
\let\thefootnote\relax\footnotetext{\textsuperscript{*}Equal contribution.}
\let\thefootnote\relax\footnotetext{\textsuperscript{\dag}Corresponding author.}
\let\thefootnote\relax\footnotetext{This work was
supported by a grant from Ford Motor Company via the Ford-UM Alliance
under award N028603.}

\input{Introduction}

\input{Related_works}

\input{Method}

\input{Experiments}

\input{Conclusion}

{
    \small
    \bibliographystyle{ieeenat_fullname}
    \bibliography{egbib}
}

\input{Supplementary}

\end{document}

%% file: Introduction.tex
\section{Introduction}



Perception is an important module for achieving safe and effective autonomous driving~\cite{carmichael2024dataset, hu2023UniAD, wilson2023convolutional}. 3D object detection is an essential task in perception as it is of great significance for subsequent tasks~\cite{wang2023_multi_modal_3d_object_detection_survey, song2024lirafusion, wilson2022motionsc, Pang_2023_standing}. 
Among various perceptual sensors used by autonomous driving researchers, LiDAR, camera and radar are the most common choices to enable autonomy on the road~\cite{wang2023_multi_modal_3d_object_detection_survey}. Sensor fusion techniques are usually used to improve the detector's performance and robustness. LiDAR-Camera (LC) fusion has been widely demonstrated as the top-performing sensor configuration for 3D object detection~\cite{nuscenes, mmdet3d2020, liu2023bevfusion, Transfusion, yang2022deepinteraction, ealss}. 
However, the high cost of LiDAR constrains the wide application of this configuration. Though Camera-Only (CO) detectors have demonstrated impressive performance in recent Bird's-Eye-View (BEV) based frameworks~\cite{li2022bevformer, li2023bevdepth, huang2021bevdet, huang2022bevdet4d}, 
the camera's vulnerability to lighting conditions and lack of accurate depth measurements motivates researchers to turn to other sensors such as radar. Radar is robust to varying weather and lighting conditions and features automotive-grade design and low cost. Radars are already highly accessible on most cars equipped with driver assistance features.
Compared with LiDAR, the radar measurements are sparse and noisy, which makes designing a Camera-Radar (CR) detector challenging. Recent CR detectors have leveraged the advancement brought by BEV-based Camera-Only (CO) detectors ~\cite{li2022bevformer, huang2021bevdet, huang2022bevdet4d, li2023bevdepth} to achieve further improvement in accuracy and robustness to weather and lighting changes~\cite{HVDetFusion, zhou2022towards_deep_radar_survey}.

\begin{figure}[t!]
\begin{center}
  \includegraphics[width=\linewidth]{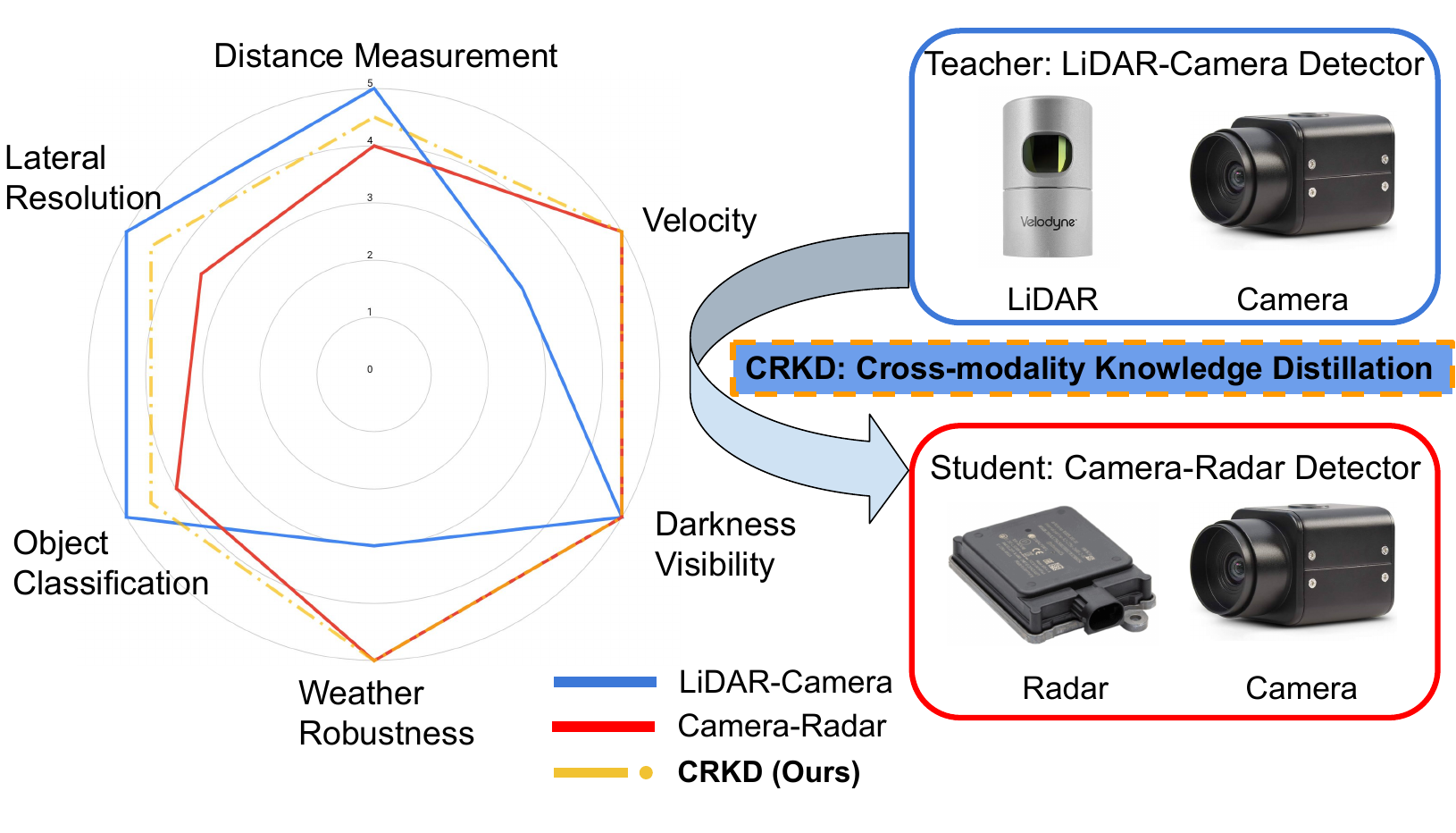}
  \caption{We propose CRKD to conduct a novel cross-modality knowledge distillation path from a LiDAR-camera teacher to a camera-radar student. We present a radar chart to illustrate the complementary nature of these sensing configurations and the improvement that CRKD can enable.}
  \label{fig:pitch}
\end{center}
\vspace{-8mm}
\end{figure}

Despite the advancement in architecture design, there is still a distinct performance gap when comparing LiDAR-Only (LO) and LC detectors against CO and CR detectors. Recent research has focused on applying the Knowledge Distillation (KD) technique to alleviate this gap~\cite{hinton_KD, zhou_UniDistill, X3KD, Wang_2023_DistillBEV, hong2022cmkd}. 
Generally, KD features a teacher-student framework that aims to propagate the informative knowledge from a well-performing teacher model to facilitate the learning process of the student model. This usually leads to improved performance compared to simply training the student model on the same task. The KD technique has been employed in either intra-modal~\cite{zhang2023pointdistiller, wei2022lidardistillation, zeng2023fd3d, li2022bevlgkd} or cross-modal~\cite{chen2022bevdistill, chong2022monodistill, hong2022cmkd, huang2022tig, zhou_UniDistill, X3KD} configurations for 3D object detection. 
Though many cross-modal methods use a single-modality detector as the teacher model to leverage the privileged LiDAR data that is widely available in open-source datasets, they mainly focus on distilling knowledge to a LiDAR-based or camera-based student detector. We argue the importance of designing a distillation path from an LC teacher detector to a CR student detector, which could benefit from the existing superior design of LC detectors and the shared point cloud representation between measurements of LiDAR and radars~\cite{RadarNet}.

Inspired by the above observations, we propose \textbf{CRKD}: an enhanced \textbf{C}amera-\textbf{R}adar 3D object detector with cross-modality \textbf{K}nowledge \textbf{D}istillation (\cref{fig:pitch}) that distills knowledge from an LC teacher detector to a CR student detector. To our best knowledge, CRKD is the first KD framework that supports a fusion-to-fusion distillation path. As the LiDAR sensor is used only during training, we emphasize the value of CRKD as it could facilitate the practical application of perceptual autonomy with a low-cost and robust CR sensor configuration.

To summarize, our main contributions are as follows:
\begin{itemize}
    \item We propose a novel cross-modality KD framework to enable LC-to-CR distillation in the BEV feature space. With the transferred knowledge from an LC teacher detector, the CR student detector can outperform existing baselines without additional cost during inference. 
    \item We design four KD modules to address the notable discrepancies between different sensors to realize effective cross-modality KD. As we operate KD in the BEV space, the proposed loss designs can be applied to other KD configurations.
    Our improvement also includes adding a gated network to the baseline model for adaptive fusion. 
    \item We conduct extensive evaluation on nuScenes~\cite{nuscenes} to demonstrate the effectiveness of CRKD. CRKD can improve the mAP and NDS of student detectors by $3.5\%$ and $3.2\%$ respectively. As our method focuses on a novel KD path with a large modality gap, we provide thorough study and analysis to support our design choices.
    
\end{itemize}

%% file: Related_works.tex
\section{Related Works}



\begin{figure*}[t!]
\begin{center}
  \includegraphics[width=0.99\linewidth]{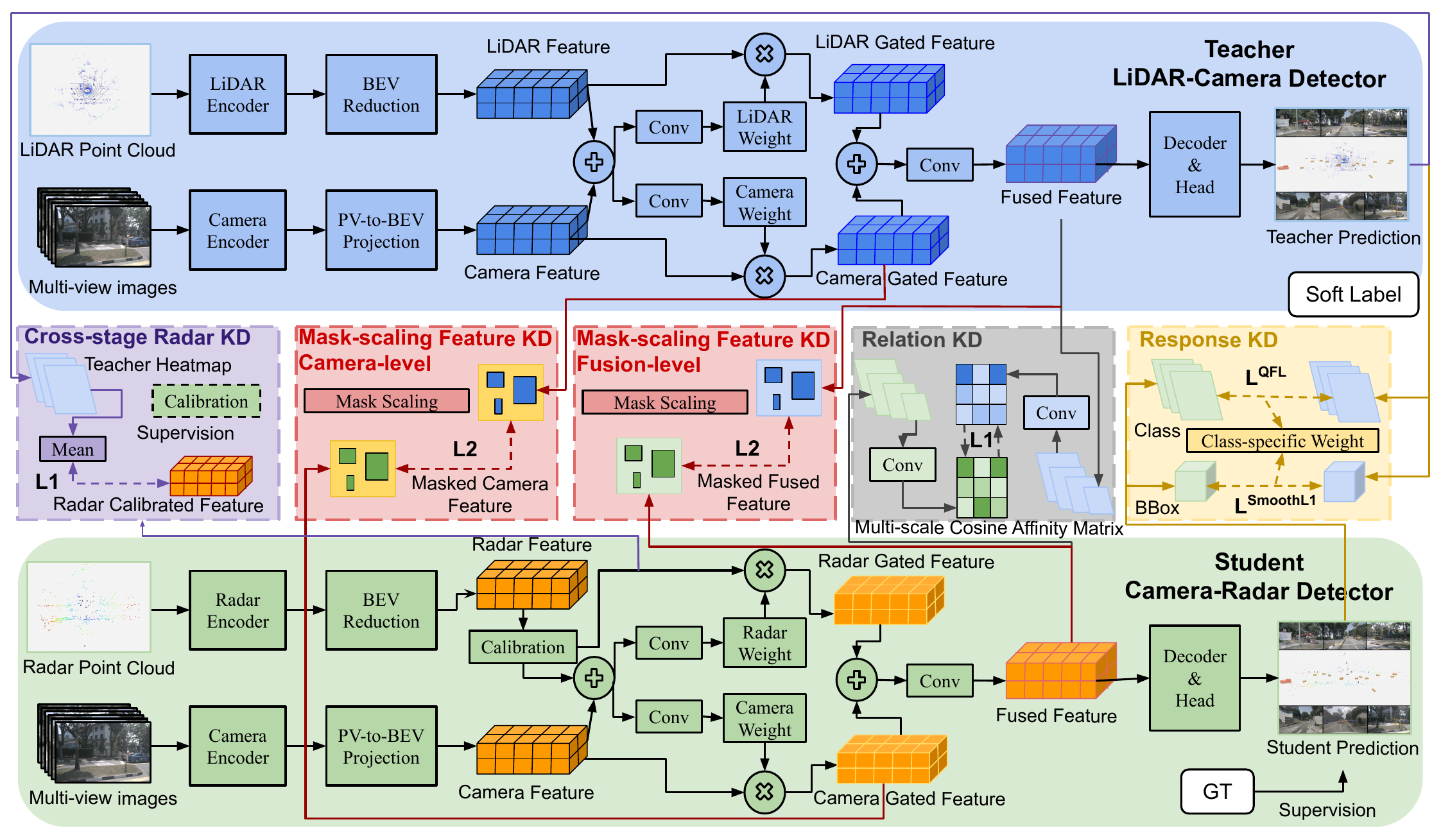}
  \caption{An overview of the proposed cross-modality (LC-to-CR) distillation framework CRKD. We narrow down the modality gap by unifying both the teacher and the student into the BEV space with similar 3D object detector structure. We refine the model design to enable adaptive fusion and design four novel distillation losses for effective cross-modality KD. During inference, only CR input is needed.}
  \label{fig:overview}
\end{center}
\vspace{-9mm}
\end{figure*}

\subsection{Multi-modality 3D Object Detection}
The multi-modality 3D object detectors generally outperform single-modality detectors in accuracy and robustness as the perceptual sensors (e.g., LiDAR, camera and radar) can complement each other~\cite{wang2023_multi_modal_3d_object_detection_survey}.
 Among the common sensor combinations, LC is the best-performing modality configuration on most existing datasets~\cite{nuscenes, geiger2012kitti, Waymo, mmdet3d2020}. In general, LiDAR and camera are fused in different ways. One trend is to augment LiDAR points with features from cameras~\cite{PointPainting, PointAugmenting, sindagi2019mvxnet, huang2020epnet, xu2021fusionpainting}, which is usually referred to as early fusion. Other solutions apply deep feature fusion in a shared representation space~\cite{3DCVF, li2022unifying_uvtr, li2022vff}. One trending choice is to leverage the BEV space to deliver impressive improvement~\cite{liu2023bevfusion, liang2022bevfusionpeking}. There are also methods that fuse information at a later stage. In~\cite{Transfusion, FUTR3D, cao2023cmt}, features are independently extracted and aggregated via proposals or queries in the detection head, while some methods~\cite{pang2020clocs, pang2022fast_clocs} combine the output candidates from single-modality detectors.

Nevertheless, the LC configuration is less accessible to consumer cars due to the high cost of LiDAR. Thus, the potential for CR detectors stands out due to the robustness brought by radars and the potential of large-scale deployment. One of the key challenges facing CR detectors is how to handle the discrepancy in sensor views and data returns. CenterFusion~\cite{CenterFusion} applies feature-level fusion by associating radar points with image features via a frustum-based method. Following this, more feature-level fusion methods are proposed in~\cite{drews2022deepfusion2, kim2020grifnet, zhou2023rcbev, wu2023mvfusion}. Recently, as many camera-based methods \cite{li2022bevformer, li2023bevdepth} have started to leverage the unified BEV space by transforming camera features from perspective view (PV), many CR detectors also explore fusion of camera and radar features in the BEV space~\cite{kim2023crn, kim2023rcmfusion}. 
Though LiDAR data is not available in real inference and deployment, its wide appearance in open-source dataset \cite{nuscenes} has motivated researchers to leverage LiDAR data to guide the feature transformation process~\cite{HVDetFusion}. Motivated by the aforementioned works, CRKD also unifies features in the BEV space and leverages LiDAR data in a KD-based framework. To our best knowledge, CRKD is the first framework that improves CR detectors with cross-modality KD from a top-performing LC teacher detector.


\subsection{Cross-modality Knowledge Distillation}
The idea of KD is initially proposed in~\cite{hinton_KD} for model compression in an image classification task. It is then extended to the field of object detection for model compression and performance improvement~\cite{KDOD_survey}. Specifically, in the field of 3D object detection, a group of KD methods requires that the teacher and student models use the same modality such as LO-to-LO (L2L)~\cite{zhang2023pointdistiller, wei2022lidardistillation, wang2021objectdgcnn, yang2022towards, ju2022paintanddistill} and CO-to-CO (C2C)~\cite{zeng2023fd3d, zhang2022structured, li2022bevlgkd}. 
In contrast, cross-modality KD focuses on KD with different modality configurations. Typical paths include LiDAR-to-Camera (L2C)~\cite{guo2021liga, liu2023stereodistill, chong2022monodistill, chen2022bevdistill, hong2022cmkd, li2022unifying_uvtr, huang2022tig} and Camera-to-LiDAR (C2L)~\cite{zhou_UniDistill, sautier2022image2lidar}. Recently, new cross-modality KD paths including a fusion-based modality have been explored. In UniDistill~\cite{zhou_UniDistill}, a universal framework that supports multiple KD paths is proposed. By unifying the features from different modalities to the shared BEV space, it supports L2C, C2L, LC-to-LO (LC2L) and LC-to-Camera (LC2C). DistillBEV~\cite{Wang_2023_DistillBEV} also supports L2C and LC2C by leveraging the shared BEV space. X3KD~\cite{X3KD} proposes a cross-modal and cross-task KD framework for L2C KD. It is evaluated on an LO-to-CR (L2CR) KD path as a supplementary task. However, it lacks specific consideration for the large domain discrepancies with radars and further experiments and analysis with the L2CR KD path.
Among the existing prior works, we observe the lack of KD methods that can support a fusion-to-fusion distillation path and handle domain differences with radars. We argue the importance of conducting distillation from an LC teacher to a CR student to leverage the shared BEV feature space of camera-based detectors and the shared point cloud representation of LiDAR and radar measurements.
According to our best knowledge, we are the first to investigate a KD framework with a fusion-to-fusion path. We demonstrate that our novel framework improves the detection performance of CR detectors comprehensively.

%% file: Method.tex
\section{Method}
We show an overview of CRKD in \cref{fig:overview}. We set up the teacher and student models with a similar BEV-based encoder-decoder head architecture. Taking advantage of the shared BEV feature space, we build CRKD based on the highly optimized BEVFusion~\cite{liu2023bevfusion} codebase. We use BEVFusion-LC as the teacher model and BEVFusion-CR as the baseline student model. The detector head in both models is set as CenterHead~\cite{yin2021centerpoint} for response KD.

To account for the challenging cross-modality fusion-to-fusion KD, we design several KD modules. We propose cross-stage radar distillation with a learning-based calibration module to enable the radar encoder to learn a more accurate scene-level object distribution. A mask-scaling feature KD is designed for feature imitation on foreground regions while accounting for inaccurate view transformation to BEV features for objects that are far from the sensor and dynamic. We apply a relation KD to maintain relation consistency in scene-level geometry. In addition, we improve the response KD design with class-specific loss weights to better leverage the CR model's ability to capture dynamic objects. Details of the proposed KD modules will be discussed in the following sections.


\subsection{Model Architecture Refinement}
\label{Sec:method_crkd_architecture}
We add a gated network~\cite{jacobs1991adaptive_moe, 3DCVF, song2024lirafusion} to BEVFusion~\cite{liu2023bevfusion} to enable the model to learn to generate attention weights on the single-modality feature maps to fuse the complementary modalities adaptively.
Specifically, the gated network learns the gated features as follows:
\begin{equation}
    \tilde{F}_{M_1} = F_{M_1}\times \sigma(\mbox{Conv}_{M_1}(\mbox{Concat}(F_{M_1}, F_{M_2}))),
\end{equation}
\begin{equation}
    \tilde{F}_{M_2} = F_{M_2}\times \sigma(\mbox{Conv}_{M_2}(\mbox{Concat}(F_{M_1}, F_{M_2}))),
\end{equation}
where $\tilde{F}_{M_1}$ and $\tilde{F}_{M_2}$ are the gated features for modality $M_1$ and $M_2$, 
$F_{M_1}$ and $F_{M_2}$ are the input feature maps to the gated network from the backbone and view transform module of modality $M_1$ and $M_2$, respectively, $\sigma$ denotes the sigmoid function, and $\mbox{Conv}_{M_1}$ and $\mbox{Conv}_{M_2}$ are two separate convolution layers for $M_1$ and $M_2$ that could learn channel-wise attention weights for the input features.
The output of the gated network is further fused by a convolutional fusion module in BEVFusion~\cite{liu2023bevfusion}. We apply the adaptive gated network to our teacher and student models to learn the relative importance between input modalities. This modification improves the detection performance of the teacher and student models, and also makes feature-based distillation more effective since the gated feature maps encode informative scene geometry from both input modalities.

In the LC teacher model, we denote the camera feature map as $F^T_c \in \mathbb{R}^{C^T_c \times H \times W}$ and the LiDAR feature map as $F^T_l \in \mathbb{R}^{C^T_l \times H \times W}$. Similarly, the camera and radar features in the student model are denoted as $F^S_c \in \mathbb{R}^{C^S_c \times H \times W}$ and  $F^S_r \in \mathbb{R}^{C^S_r \times H \times W}$, respectively. We denote the fused feature map in the teacher model and the student model as $F^T_f$ and $F^S_f$. We keep the spatial dimension $H\times W$ consistent for all feature maps, and also set $C^T_c = C^S_c$ and $C^T_l = C^S_l$ for feature mimicking across the feature dimension.



\subsection{Cross-Stage Radar Distillation (CSRD)}
\label{Sec:method_csrd}
Though the measurements of radar and LiDAR are both represented as point clouds, the physical meaning behind them are slightly different.
Compared with LiDAR, radar points are much sparser and can be interpreted as a list of object-level points with velocity measurements~\cite{wang2023_multi_modal_3d_object_detection_survey, RadarNet}, while LiDAR is denser and captures geometry-level information. Observing this gap, we argue the common method of direct feature imitation may not work well in this scenario. Instead, as radar measurements are sparse and represent scene-level object distribution, we propose a novel Cross-Stage Radar Distillation (CSRD) method. Specifically, we design a distillation path between the radar feature map and the scene-level objectness heatmap predicted by the LC teacher model, which is denoted as $Y^T \in \mathbb{R}^{K \times H \times W}$, where $K$ is the number of classes. Since radars are generally believed to be noisy in the range and azimuth angle measurements, we design a calibration module to learn to compensate the noise. Specifically, we pass $F^S_r$ to three blocks of convolutional layer, batch normalization and ReLU activation with kernel size $3\times3$. We add another $1\times1$ convolution layer to project the calibrated feature map to $\hat{F}^S_r \in \mathbb{R}^{H \times W}$. The CSRD loss $\mathcal{L}_{csrd}$ is formed as follows:
\newcommand\mathbox[1]{\mbox{$#1$}}
\begin{equation}
    \mathcal{L}_{csrd} = \frac{1}{H\times W}\sum _{i}^{H}\sum _{j}^{W} \Vert \hat{Y}^T_{i,j} - \mathbox{\hat{F}^S_r}_{i,j}\Vert_1,
\end{equation}
where $\hat{Y}^T \in \mathbb{R}^{H \times W}$ is obtained by taking the mean along the $K$ dimension of $\sigma(Y^T)$.

\subsection{Mask-Scaling Feature Distillation (MSFD)}
We propose feature distillation for aligning the camera feature maps and the fused feature maps.
It has been acknowledged in many works~\cite{zhou_UniDistill, chong2022monodistill, chen2022bevdistill, zhao2023bevsimdet} that direct feature imitation between teacher and student models may not work effectively in 3D object detection tasks due to the notable imbalance between the foreground and background. Therefore, a common fix to this issue is to generate a mask $M \in \mathbb{R}^{H\times W}$ to only distill information from the foreground region. Meanwhile, more works have demonstrated that the boundary region of the foreground can also contribute to effective KD~\cite{chen2022bevdistill, zhao2023bevsimdet}. We follow this finding and propose Mask-Scaling Feature Distillation (MSFD) that is aware of object range and movement. For the student CR model, the detection performance is mainly dependent on the depth prediction for images and the geometric accuracies of radar points. Since the range and object movement can cause extra challenges for view transformation to BEV space, we scale up the area of the foreground region to account for the potential misalignment.
We increase the width and length of the mask by $\alpha$ and $\beta$ if the objects are in the range groups $[r_1, r_2]$ and $[r_2, \infty]$. We also increase the width and length by $\alpha$ and $\beta$ if the velocities along that axis are within $[v_1, v_2]$ and $[v_2, \infty]$. In practice, 
We clip the increase of object size within a pre-defined range to balance between different sizes of objects. We form the MSFD loss as follows:
\begin{equation}
    \mathcal{L}_{msfd} = \frac{1}{H\times W}\sum _{i}^{H}\sum _{j}^{W} M_{i,j} \Vert{F_{i,j}^{T}} -F_{i,j}^{S}\Vert_2,
\end{equation}
where $F^T$ and $F^S$ represent the feature maps in the teacher and student model, respectively. We compute MSFD loss for the gated camera feature maps ($\tilde{F}^T_c$ and $\tilde{F}^S_c$) and the fused feature maps ($F^T_f$ and $F^S_f$).

\subsection{Relation Distillation (RelD)}
While the aforementioned CSRD and MSFD can handle feature-level distillation effectively, we follow MonoDistill~\cite{chong2022monodistill} to highlight the importance of maintaining similar geometric relations in the scene level between the teacher and student models. We compute the affinity matrix describing cosine similarity of the fused feature map. We propose the RelD loss as follows:

\begin{equation}
    C_{i, j} = \frac{F_{i}^\top F_{j}}{\Vert F_{i} \Vert_2 \cdot \Vert F_{j} \Vert_2},
\end{equation}
where $C_{i, j}$ denotes the cosine similarity value at $(i, j)$ in the affinity matrix, and $F_{i}$ and $F_{j}$ represent the $i^{th}$ and $j^{th}$ feature map separately. The scene-level information gap between the student and teacher models can be computed as the $\mathcal{L}_{1}$ norm between their respective affinity matrices. We refer to the RelD loss as $\mathcal{L}_{reld}$, as shown below:
\begin{equation}
    \mathcal{L}_{reld} = \frac{1}{H \times W}\sum\limits_{i=1}^{H}\sum\limits_{j=1}^{W}\Vert C^{T}_{i, j} - C^{S}_{i, j} \Vert_1,
\end{equation}
where $H$ and $W$ represent the BEV spatial size, and $C^{T}$ and $C^{S}$ denote the affinity matrix of the student and teacher model, respectively.
In CRKD, we compute $\mathcal{L}_{reld}$ between the fused feature maps of the teacher and student models since they are the input to the decoder and detector heads. The refined feature maps with distilled relation information could improve the detection performance.
Moreover, in order to distill the scene-level relation information of different scales, we apply a downsampling operation and a convolutional block. Then we use these multi-level feature maps to calculate the multi-scale RelD losses and take the average value as the final loss term.

\subsection{Response Distillation (RespD)}
    Response Distillation has been proven effective in image classification \cite{hinton_KD} and 3D object detection \cite{hong2022cmkd, zhou_UniDistill, Wang_2023_DistillBEV}. The predictions inferred by the teacher are served as the soft labels for the student. The soft labels and the hard labels are combined to supervise the learning of the student model. We refer to the RespD design in CMKD~\cite{hong2022cmkd} and improve it to be aware of modality strength. Since radar has the unique advantage of direct velocity measurements due to the Doppler effect~\cite{zhou2022towards_deep_radar_survey, RadarNet}, we set larger weights for dynamic classes in RespD to allow higher priority for dynamic objects to leverage the student CR model's strength.
The loss for dynamic response distillation is denoted as $\mathcal{L}_{resp}$, consisting of the classification loss $\mathcal{L}_{cls}$ and the regression loss $\mathcal{L}_{reg}$.
$\mathcal{L}_{cls}$ is for object categories and is computed using the Quality Focal Loss (QFL) \cite{li2020qfl}.
$\mathcal{L}_{reg}$ is for 3D bounding box regression and can be obtained by calculating the $SmoothL1$ loss.
We compute these two losses as:
\begin{equation}
    \mathcal{L}_{cls} = \sum\limits_{i=1}^{K}
    QFL(P^{T}_{C_{i}}, P^{S}_{C_{i}}) \times w_{i},
\end{equation}
\begin{equation}
    \mathcal{L}_{reg} = \sum\limits_{i=1}^{K}
    Smooth\mathcal{L}1(P^{T}_{B_{i}}, P^{S}_{B_{i}}) \times w_{i},
\end{equation}
where $P^{T}_{C_{i}}$ and $P^{S}_{C_{i}}$ denote the classification predictions of the $i^{th}$ task generated by the teacher and student model, and $P^{T}_{B_{i}}$ and $P^{S}_{B_{i}}$ denote the regression predictions in the teacher and student models. $K$ is the number of tasks in the Centerhead~\cite{yin2021centerpoint}, and $w_{i}$ represents the class-specific weights.


\subsection{Overall Loss Function}
We combine the proposed KD loss and standard 3D object detection loss $\mathcal{L}_{det}$. The overall loss function we apply in the training stage is: 

\begin{equation}
    \mathcal{L} = \lambda_{1} \cdot \mathcal{L}_{csrd} + \lambda_{2} \cdot \mathcal{L}_{msfd} + \lambda_{3} \cdot \mathcal{L}_{reld} + \lambda_{4} \cdot \mathcal{L}_{respd} + \lambda_{5} \cdot \mathcal{L}_{det},
\end{equation}

\noindent where $\lambda_1$ through $\lambda_5$ are hyperparameters we set for weighting different loss components.

%% file: Experiments.tex
\section{Experiments}

\begin{table*}[t!]
  \centering
  
  \resizebox{\linewidth}{!}{
  \begin{tabular}{c|c|c|c|c|ccccccc}
    \toprule
    Set & Method & Modality & Backbone & Resolution & mAP$\uparrow$ & NDS$\uparrow$ & mATE$\downarrow$ & mASE$\downarrow$ & mAOE$\downarrow$ & mAVE$\downarrow$ & mAAE$\downarrow$  \\
    \midrule
    ~ & BEVFormer-S~\cite{li2022bevformer} & C & R101 & $900 \times 1600$ & 37.5 & 44.8 & 0.725 & 0.272 & \textbf{0.391} & 0.802 & 0.200\\
    ~ & BEVDet~\cite{huang2021bevdet} & C & R50 & $256 \times 704$ & 29.8 & 37.9 & 0.725 & 0.279 & 0.589 & 0.860 & 0.245\\
    \cmidrule(lr){2-12}
    ~ & RCM-Fusion~\cite{kim2023rcmfusion} & C+R & R101 & $900 \times 1600$ & \underline{44.3} & 52.9 & - & - & - & - & -\\
    ~ & CenterFusion~\cite{CenterFusion} &  C+R & DLA34 & $450 \times 800$ & 33.2 & 45.3 & 0.649 & \textbf{0.263} & 0.535 & 0.540 & \textbf{0.142}\\
    ~ & CRAFT~\cite{kim2023craft} & C+R & DLA34 & $448 \times 800$ & 41.1 & 51.7 & 0.494 & 0.276 & 0.454 & 0.486 & 0.176\\
    ~ & RCBEV~\cite{zhou2023rcbev} & C+R & SwinT & $256 \times 704$ & 37.7 & 48.2 & 0.534 & 0.271 & 0.558 & 0.493 & 0.209\\
    ~ & BEVFusion~\cite{liu2023bevfusion} & C+R & SwinT & $256 \times 704$ & 43.2 & 54.1 & 0.489 & 0.269 & 0.512 & \textbf{0.313} & 0.171\\
    \cmidrule(lr){2-12}
    ~ & UVTR (L2C)~\cite{li2022unifying_uvtr} & C$\Diamond$ & R101 & $900 \times 1600$ & 37.2 & 45.0 & 0.735 & 0.269 & \underline{0.397} & 0.761 & 0.193\\
    ~ & BEVDistill (BEVFormer-S)~\cite{chen2022bevdistill} & C$\Diamond$ & R50 & $640 \times 1600$ & 38.6 & 45.7 & 0.693 & \underline{0.264} & 0.399 & 0.802 & 0.199\\
    ~ & UniDistill (LC2C)~\cite{zhou_UniDistill} & C$\Diamond$ & R50 & $256 \times 704$ & 26.5 & 37.8 & - & - & - & - & -\\
    ~ & BEVSimDet~\cite{zhao2023bevsimdet} & C$\Diamond$ & SwinT & $256 \times 704$ & 40.4 & 45.3 & 0.526 & 0.275 & 0.607 & 0.805 & 0.273\\
    ~ & X3KD (LC2C)~\cite{X3KD} & C$\Diamond$ & R50 & $256 \times 704$ & 39.0 & 50.5 & 0.615 & 0.269 & 0.471 & 0.345 & 0.203\\
    ~ & DistillBEV (BEVDet)~\cite{Wang_2023_DistillBEV} & C$\Diamond$ & R50 & $256 \times 704$ & 34.0 & 41.6 & 0.704 & 0.266 & 0.556 & 0.815 & 0.201\\
    \cmidrule(lr){2-12}
    ~ & X3KD (L2CR)~\cite{X3KD} & C+R$\Diamond$ & R50 & $256 \times 704$ & 42.3 & 53.8 & - & - & - & - & -\\
    \rowcolor{black!10}
    \cellcolor{white} 
     & \textbf{CRKD} & C+R$\Diamond$ & R50 & $256 \times 704$ & 43.2 & \underline{54.9} & \underline{0.450} & 0.267 & 0.442 & 0.339 & 0.176\\
    \rowcolor{black!10}
    \cellcolor{white} 
    \multirow{-16}*{\emph{val}} & \textbf{CRKD} & C+R$\Diamond$ & SwinT & $256 \times 704$ & \textbf{46.7} & \textbf{57.3} & \textbf{0.446} & \textbf{0.263} & 0.408 & \underline{0.331} & \underline{0.162}\\

    \midrule
    ~ & BEVFormer-S~\cite{li2022bevformer} & C & R101 & $900 \times 1600$ & 40.9 & 46.2 & 0.650 & 0.261 & 0.439 & 0.925 & 0.147\\
    ~ & BEVDet$\dagger$~\cite{huang2021bevdet} & C & SwinT & $640 \times 1600$ & 42.4 & 48.2 & 0.528 & \textbf{0.236} & 0.395 & 0.979 & 0.152\\

    \cmidrule(lr){2-12}
    
    ~ & RCM-Fusion~\cite{kim2023rcmfusion} & C+R & R101 & $900 \times 1600$ & \textbf{49.3} & \underline{58.0} & 0.485 & 0.255 & \underline{0.386} & 0.421 & 0.115\\ 
    ~ & CenterFusion$\dagger$~\cite{CenterFusion} & C+R & DLA34 & $450 \times 800$ & 32.6 & 44.9 & 0.631 & 0.261 & 0.516 & 0.614 & 0.115\\ 
    ~ & CRAFT$\dagger$~\cite{kim2023craft} & C+R & DLA34 & $448 \times 800$ & 41.1 & 52.3 & \underline{0.467} & 0.268 & 0.456 & 0.519 & \underline{0.114}\\ 
    ~ & RCBEV~\cite{zhou2023rcbev} & C+R & SwinT & $256 \times 704$ & 40.6 & 48.6 & 0.484 & 0.257 & 0.587 & 0.702 & 0.140\\

    \cmidrule(lr){2-12}
    
    ~ & UVTR (L2C)~\cite{li2022unifying_uvtr} & C$\Diamond$ & V2-99 & $900 \times 1600$ & 45.2 & 52.2 & 0.612 & 0.256 & \textbf{0.385} & 0.664 & 0.125\\
    ~ & X3KD (LC2C)~\cite{X3KD} & C$\Diamond$ & R101 & $640 \times 1600$ & 45.6 & 56.1 & 0.506 & \underline{0.253} & 0.414 & \textbf{0.366} & 0.131\\ 
    ~ & UniDistill (LC2C)~\cite{zhou_UniDistill} & C$\Diamond$ & R50 &  $256 \times 704$ & 29.6 & 39.3 & 0.637 & 0.257 & 0.492 & 1.084 & 0.167\\ 
    \cmidrule(lr){2-12}
    ~ & X3KD (L2CR)~\cite{X3KD} & C+R$\Diamond$ & R50 & $256 \times 704$ & 44.1 & 55.3 & - & - & - & - & -\\
    \rowcolor{black!10}
    \cellcolor{white} 
    \multirow{-11}*{\emph{test}} & \textbf{CRKD} & C+R$\Diamond$ & SwinT & $256 \times 704$ & \underline{48.7} & \textbf{58.7} & \textbf{0.404} & \underline{0.253} & 0.425 & \underline{0.376} & \textbf{0.111}\\
    \bottomrule
  \end{tabular}
  
  }
  \caption{Comparison on the nuScenes dataset. We group methods based on modality and whether KD is used. We include existing SOTA works that use single-frame image input for fair comparison. Methods~\cite{chen2022bevdistill, Wang_2023_DistillBEV, zhao2023bevsimdet} missed in the \emph{test} split group do not report their results with single-frame image input. The proposed CRKD outperforms the baseline methods in most metrics. $\Diamond$ denotes the distillation-based methods. $\dagger$ denotes using test time augmentation. The best is \textbf{bolded} and the second best is \underline{underlined}.}
  \label{tab:quantitative}
  \vspace{-5mm}
\end{table*}

\subsection{Experimental Setup}
We evaluate our method on the nuScenes dataset \cite{nuscenes} as the three modalities (i.e., LiDAR, camera and radar) are all available. We follow the official split that has 700 scenes for training and 150 scenes for validation. We set $[-54m, -54m, -5m] \times [54m, 54m, 3m]$ as the region to conduct object detection. We use the mean-Average-Precision (mAP) and NuScenes Detection Score (NDS)~\cite{nuscenes} as the main evaluation metrics. We also report the True Positive (TP) metrics~\cite{nuscenes} for comprehensive evaluation.

Our implementation is based on the MMDetection3D codebase~\cite{mmdet3d2020}. All of our experiments are conducted using 4 NVIDIA A100 GPUs. 
We set the default BEVFusion-LC with Centerhead~\cite{yin2021centerpoint} as the teacher model. As mentioned in \cref{Sec:method_crkd_architecture}, we add the adaptive gated network to the baseline BEVFusion-CR model and denote it as BEVFusion-CR*. We set BEVFusion-CR* as the student model. We set the camera backbone as SwinT~\cite{liu2021swin} and image resolution as $256 \times 704$ for both the teacher and student models. We also test CRKD with a ResNet R50 backbone~\cite{He_2016_resnet} for comprehensive evaluation. The BEV spatial size is set to $180 \times 180$. We use PointPillars~\cite{lang2019pointpillars} as the backbone of the radar branch in BEVFusion-CR*. We set the same Centerhead~\cite{yin2021centerpoint} as the detector head for the student model. During distillation,  we freeze the teacher model and train the student model for 20 epochs. We set the batch size as $8$ and learning rate as 1e-4. We include more implementation details and results in the supplementary material.


\subsection{Quantitative Results}
We give an overall comparison of CRKD with existing CO and CR detectors with single-frame image input on nuScenes~\cite{nuscenes}. We follow common practices~\cite{chen2022bevdistill, zhou_UniDistill, X3KD, Wang_2023_DistillBEV} to show a complete comparison on both the \emph{val} and \emph{test} splits. As shown in~\cref{tab:quantitative}, CRKD is the top-performing model in most metrics on the \emph{val} set of nuScenes. We also present the performance of CRKD on the \emph{test} split. The results show that CRKD has the best or second best performance on most metrics without using any test-time optimization techniques (e.g., test-time augmentation, larger image resolution). Overall, CRKD is the most consistent in achieving high performance across all of the baselines.

We also show a complete comparison of per-class AP in \cref{tab:perclass} to break down the improvement brought by CRKD. The results show that CRKD achieves consistent improvement in AP of all the classes. There is an interesting finding that larger gains of CRKD come from dynamic classes, which indicates that CRKD successfully helps the student model leverage its strength on dynamic object detection more effectively due to the availability of direct velocity measurements from radar.

\begin{table*}[!h]
  \centering
    \resizebox{\linewidth}{!}{
  \begin{tabular}{c|c|cccccccccc|c}
    \toprule
    Model & Modality & Car & Truck & Bus & Trailer & CV & Ped & Motor & Bicycle & TC & Barrier & mAP$\uparrow$\\
    \midrule
    Teacher & L+C & 88.4 & 62.4 & 73.8 & 40.6 & 29.2 & 78.7 & 75.3 & 65.8 & 74.9 & 72.3 & 66.1\\
    \midrule
    Baseline & C+R & 72.1 & 37.8 & 48.9 & 18.3 & 12.6 & 48.4 & 42.0 & 33.8 & 58.8 & 59.6 & 43.2\\
    Student & C+R & 72.2 & 41.3 & 51.0 & 19.2 & 15.2 & 49.0 & 46.2 & 35.5 & 59.1 & \textbf{60.1} & 44.9\\
    \rowcolor{black!10}
    \cellcolor{white} 
    \textbf{CRKD} & C+R & \textbf{74.8(+2.7)} & \textbf{44.1(+6.3)} & \textbf{53.6(+4.7)} & \textbf{20.6(+2.3)} & \textbf{16.9(+4.3)} & \textbf{50.6(+2.2)} & \textbf{46.8(+4.8)} & \textbf{38.2(+4.4)} & \textbf{61.5(+2.7)} & \textbf{60.1(+0.5)} & \textbf{46.7(+3.5)}\\
    \bottomrule
  \end{tabular}
}
  \caption{Comparison of the per-class AP results of the BEVFusion-LC (teacher), BEVFusion-CR (baseline), BEVFusion-CR* (student) and CRKD models on the nuScenes \emph{val} split. We quantitatively show the improvement made by CRKD over the baseline.}
  \label{tab:perclass}
  \vspace{-2mm}
\end{table*}

\begin{table*}[h]
  \centering
  \resizebox{\linewidth}{!}{
  \begin{tabular}{c|ccccc|ccccccc}
    \toprule
    Model & Gated & RespD & CSRD & MSFD & RelD & mAP$\uparrow$ & NDS$\uparrow$ & mATE$\downarrow$ & mASE$\downarrow$ & mAOE$\downarrow$ & mAVE$\downarrow$ & mAAE$\downarrow$\\
    \midrule
    Baseline & & & & & & 43.2 & 54.1 & 0.489 & 0.269 & 0.512 & 0.313 & 0.171\\
    \midrule
    \rowcolor{black!2}
    \cellcolor{white} ~ & \checkmark & & & & & 44.9 & 55.9 & 0.464 & 0.267 & 0.458 & 0.304 & 0.165\\
    \rowcolor{black!6}
    \cellcolor{white} ~ & \checkmark & \checkmark & & & & 45.7 & 56.7 & 0.448 & 0.262 & 0.409 & 0.330 & 0.166\\
    \rowcolor{black!10}
    \cellcolor{white} ~ & \checkmark & \checkmark & \checkmark & & & 46.0 & 57.0 & 0.445 & 0.261 & 0.407 & 0.326 & 0.163\\
    \rowcolor{black!14}
    \cellcolor{white} ~ & \checkmark & \checkmark & \checkmark & \checkmark & & 46.2 & 57.2 & 0.439 & 0.260 & 0.394 & 0.332 & 0.166\\
    \rowcolor{black!18}
    \cellcolor{white} \multirow{-4}*{CRKD} & \checkmark & \checkmark & \checkmark & \checkmark & \checkmark & 46.7 & 57.3 & 0.446 & 0.263 & 0.408 & 0.331 & 0.162\\
    \bottomrule
  \end{tabular}
  }
  \caption{Ablation study of the proposed modules in CRKD evaluated on the nuScenes \emph{val} split. The baseline is the BEVFusion-CR model.}
  \label{tab:main_ablation}
    \vspace{-2mm}
\end{table*}

\begin{table*}[!htp]
    \centering
    \begin{subtable}{0.49\linewidth}
        \centering
      \begin{tabular}{c|cc|cc}
        \toprule
        Module & LiDAR & Heatmap & mAP$\uparrow$ & NDS$\uparrow$ \\
        \midrule
        ~ & \checkmark &  &44.9  &56.3 \\
        \rowcolor{black!10}
        \cellcolor{white} 
        \multirow{-2}*{CSRD} &  & \checkmark & 46.0 & 57.0\\
        \bottomrule
      \end{tabular}
      \caption{Ablation study of Cross-stage Radar Distillation (CSRD).}
      \label{tab:csrd}
    \end{subtable}
    \begin{subtable}{0.49\linewidth}
        \centering
      \begin{tabular}{c|cc|cc}
        \toprule
        Module & Mask & Mask-scaling & mAP$\uparrow$ & NDS$\uparrow$ \\
        \midrule
        ~ & \checkmark &  & 46.0 & 56.7\\
        \rowcolor{black!10}
        \cellcolor{white} 
        \multirow{-2}*{MSFD} &  & \checkmark & 46.2 & 56.9\\
        \bottomrule
      \end{tabular}
      \caption{Ablation study of Mask-scaling Feature Distillation (MSFD).}
      \label{tab:msfd}
    \end{subtable}
    \begin{subtable}{0.49\linewidth}
        \centering
      \begin{tabular}{c|cc|cc}
        \toprule
        Module & Vanilla & Adapt & mAP$\uparrow$ & NDS$\uparrow$ \\
        \midrule
        ~ & \checkmark &  & 45.9 & 56.9\\
        \rowcolor{black!10}
        \cellcolor{white} \multirow{-2}*{RelD} &  & \checkmark & 46.2 & 57.0\\
        \bottomrule
      \end{tabular}
      \caption{Ablation study of Relation Distillation (RelD).}
      \label{tab:reld}
    \end{subtable}
    \begin{subtable}{0.49\linewidth}
        \centering
      \begin{tabular}{c|cc|cc}
        \toprule
        Module & Vanilla & Dynamic & mAP$\uparrow$ & NDS$\uparrow$ \\
        \midrule
        ~ & \checkmark &  & 45.3 & 56.7 \\
        \rowcolor{black!10}
        \cellcolor{white} \multirow{-2}*{RespD} &  & \checkmark & 45.7 & 56.7\\
        \bottomrule
      \end{tabular}
      \caption{Ablation study of Response Distillation (RespD).}
      \label{tab:respd}
    \end{subtable}

\caption{Ablation Study of single distillation modules on the nuScenes \textit{val} split.}
\label{tab:module_ablation}
  \vspace{-4mm}
\end{table*}

\begin{table}[!h]
  \centering
  \begin{tabular}{c|c|c|cc}
    \toprule
    \textit{Fuser} & \textit{In Channels} & \textit{Out Channels} & \textit{mAP$\uparrow$} & \textit{NDS$\uparrow$} \\
    \midrule
    Conv & 80 + 256 & 256 & 43.2 & 54.1\\
    \midrule
    ~ & 64 + 64 & 64 & 44.2 & 54.3\\
    ~ & 128 + 128 & 256 & 44.4 & 54.7\\
    \rowcolor{black!10}
    \cellcolor{white} \multirow{-3}*{Gated} & 80 + 256 & 256 & 44.9 & 55.9\\
    \bottomrule
  \end{tabular}
  \caption{Ablation study of fusion module design and number of channels of the student model on the nuScenes \emph{val} split.}
  \label{tab:gated}
    \vspace{-6mm}
\end{table}

\begin{figure*}[ht!]
\begin{center}
  \includegraphics[width=0.95\linewidth]{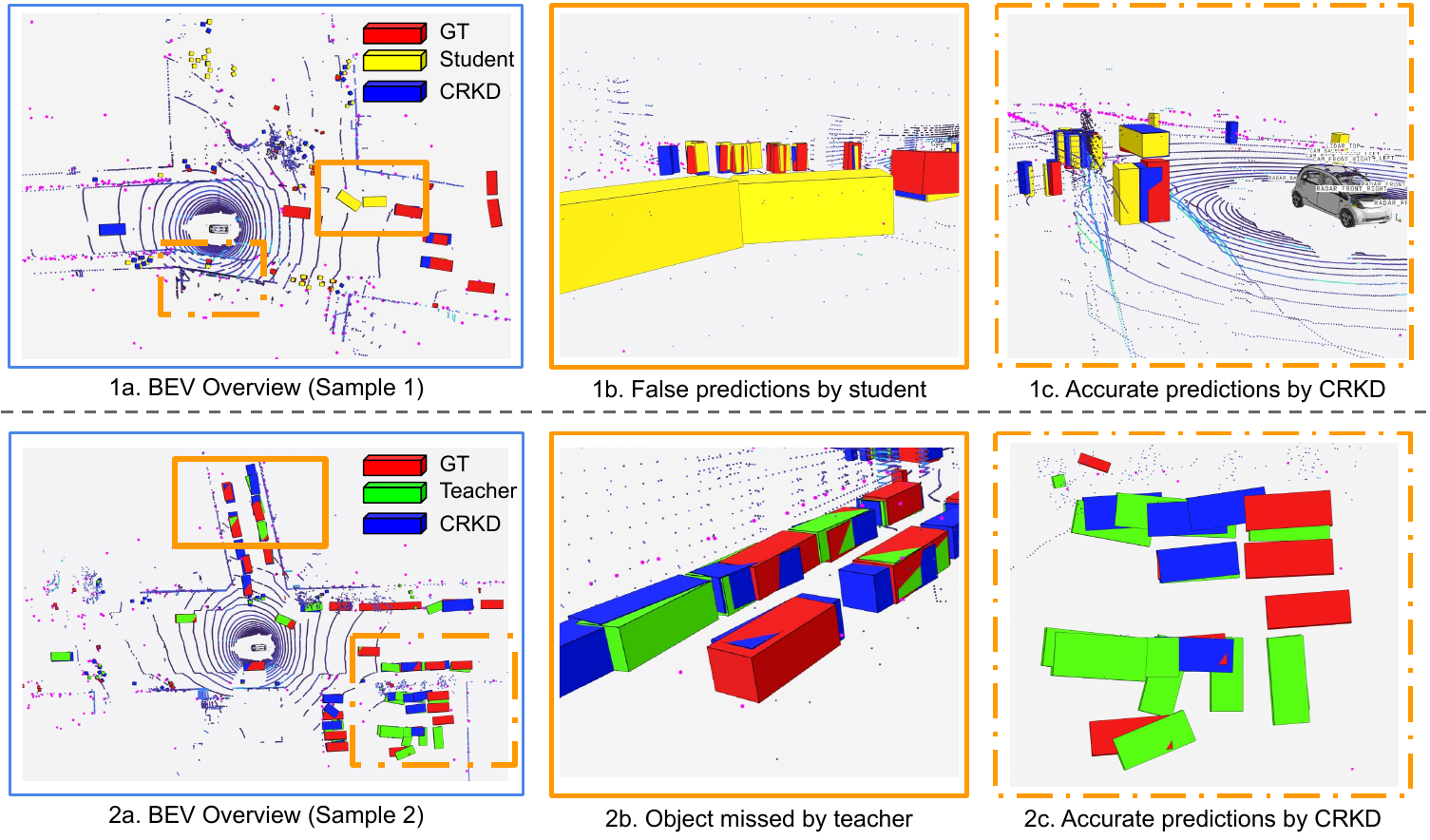}
  \caption{Qualitative results on nuScenes. We show zoomed-in views in panel b and c for the highlighted regions in panel a, with the border dash as the correspondence. We show the ground truth annotation in {\color{red}red}, teacher prediction in {\color{green}green}, student prediction in {\color{yellow}yellow}, CRKD prediction in {\color{blue}blue}, and radar points in {\color{magenta}magenta}. In (1a) to (1c), we show an example frame where CRKD has more accurate predictions and fewer false predictions than the student model. In (2a) to (2c) we show another example frame where CRKD even outperforms the LC teacher by detecting a missed car and rejecting several false predictions. Best viewed on screen and in color.}
  \label{fig:qualitative}
\end{center}
\vspace{-6mm}
\end{figure*}

\subsection{Ablation Studies}
To further break down the improvement brought from each module we design, extensive experiments are conducted to discuss and validate our design choices. We firstly present the main ablation study in~\cref{tab:main_ablation}. It can be observed that all the proposed modules have contributed to the superior performance of CRKD. Among the four proposed KD losses, we see the most improvement comes from RespD, which indicates the significance of RespD in cross-modality KD. The other three losses contribute more to the improvement of mAP. This finding validates our design objectives in improving object localization as CSRD and MSFD supervise on the feature maps, and RelD tries to align the scene-level geometric relation.

We next demonstrate the experiments we conduct to validate the design choices of each module. As mentioned before, response distillation (RespD) brings the most improvement among all the proposed KD modules. Our empirical finding indicates that for other KD modules, the best-performing design alone may not be the best choice that works with RespD and other modules. We conjecture that this inconsistency in performance gain comes from the considerably large domain discrepancies between LC and CR. We would like to highlight the importance of RespD in the cross-modality KD work and use the experiments of combining different modules with RespD to guide the overall design of CRKD. Therefore, for the following ablation study in~\cref{tab:module_ablation}, unless otherwise mentioned, we show results of experiments with using RespD together. All the ablation study instances are trained using the same setting as the full model.

\subsubsection{Effect of CSRD}
We conduct an ablation study to validate the proposed CSRD module. As mentioned in~\cref{Sec:method_csrd}, the radar points represent object-level information. Therefore, the common practice to distill information at the same stage of the network may not work well for radar feature maps. We propose CSRD to add cross-stage supervision on the object-level information. We demonstrate the model with CSRD outperforms that using LiDAR feature maps as the distillation source in~\cref{tab:csrd}. The significant improvement from CSRD validates that the higher-level objectness heatmap provides more suitable guidance to distill radar features.


\subsubsection{Effect of MSFD}
We conduct a comparison between the proposed mask-scaling strategy accounting for object range and velocity and the common foreground mask of ground truth bounding boxes. In~\cref{tab:msfd}, the improvement achieved with the proposed mask-scaling strategy verifies that our mask-scaling strategy in MSFD helps to achieve more effective KD.

\subsubsection{Effect of RelD}
We study the effect of applying a convolutional block after the downsampling operation for the feature maps used in RelD. We name the instance with the convolution layer as Adapt and the baseline instance as Vanilla in~\cref{tab:reld}. The results verify our design choice.

\subsubsection{Effect of RespD}
Though RespD has been a widely used loss term in several KD works, we are the first to weight the RespD loss differently based on different classes. In practice, we set $w_i$ as 2 for the dynamic classes and $1$ for static classes. This design allows the training supervision to prioritize dynamic classes, which radars are more capable of detecting. In the vanilla setting we set $w_i$ as $1$ for all classes. As shown in~\cref{tab:respd}, applying the class-specific weights helps to improve the overall performance of the student detector.

\subsubsection{Effect of Model Architecture Refinement}
We study the effect of adding the adaptive gated network to the original fusion module in BEVFusion~\cite{liu2023bevfusion} and tuning the number of channel dimensions. As shown in~\cref{tab:gated}, the addition of the gated network brings notable improvement over the default fusion module in the baseline BEVFusion-CR model. We also notice that matching the input and the output channel dimension of the fusion module to the teacher model (Camera: 80, LiDAR: 256) brings additional improvement. The following KD operation also gets benefit from the same channel dimension setting between the teacher and student models since no additional channel-wise projection is needed for feature-level KD.

\subsection{Qualitative Results}
We show the visualization of the 3D object detection results to highlight the effectiveness of CRKD in~\cref{fig:qualitative}. With CRKD, the detector is able to predict fewer false positive predictions and localize the objects better. We also show an impressive comparison between the teacher LC model and CRKD to demonstrate that CRKD can even outperform the teacher model with the help of radar measurements. This qualitative example validates the effectiveness of the CRKD framework and the value of radars for modern perception for autonomous driving. We will include more qualitative examples and discussion in the supplementary material.

%% file: Conclusion.tex
\section{Conclusion}
We have proposed CRKD, a novel KD framework that supports a cross-modality fusion-to-fusion KD path for 3D object detection. We leverage the BEV space to design a novel LC-to-CR KD framework. We design four distillation losses to address the significant domain gap and facilitate the distillation process in this cross-modality setting. We also introduce the adaptive gated network to learn the relative importance between two expert feature maps. Extensive experiments show the effectiveness of CRKD in improving the detection performance of CR detectors. We hope CRKD will inspire future research to leverage our proposed KD framework to further explore the potential of CR detectors to improve the reliability of this widely accessible sensor suite. In future work, we plan to extend the proposed CRKD framework to other perception tasks such as occupancy mapping.

%% file: Supplementary.tex
\maketitlesupplementary

\appendix

We provide this supplementary material with additional details to support the main paper.

\input{Implementation}

\input{Complementary_experiments}

%% file: Implementation.tex
\section{Implementation Details}
\label{sec:implementation}
In this section, we provide implementation details of the CRKD framework to enable cross-modality Knowledge Distillation (KD) from a LiDAR-Camera (LC) teacher detector to a Camera-Radar (CR) student detector.

\subsection{Data Augmentation}
 Our data processing pipeline is mainly adopted from the open-source implementation of BEVFusion~\cite{liu2023bevfusion}. The pipeline of processing the camera image is the same in the teacher and student models. The images of the cameras from 6 Perspective View (PV) are loaded and resized to $256 \times 704$. During the training process, data augmentation is applied to the images. We resize the image with scaling factors in the range of $[0.38, 0.55]$. We also set the random rotation in the range of $[-5.4 \degree, 5.4 \degree]$. The images are normalized following the default practice in~\cite{mmdet3d2020}. For the LiDAR input, the keyframe point cloud is loaded along with $9$ previous sweeps. During training, random resizing is applied with the scaling factor in $[-0.9,1.1]$. Translation augmentation is also applied with a limit of $0.5\text{m}$. For the radar input, the keyframe is loaded with $6$ previous sweeps. We follow BEVFusion~\cite{liu2023bevfusion} to select radar data dimensions. The training time augmentation of radar points is the same as LiDAR data for consistency. We also apply the class-balanced grouping and sampling (CBGS) strategy during training~\cite{zhu2019cbgs}. We do not apply any test-time augmentation for any of our models.



\subsection{Teacher Model}
As mentioned in the main paper, we add a gated network to the BEVFusion-LC~\cite{liu2023bevfusion} and denote it as BEVFusion-LC*. We use the CenterHead~\cite{yin2021centerpoint} as the detector head in the BEVFusion-LC*. There are two streams in the teacher model for LiDAR and cameras. The LiDAR point cloud is encoded as the Bird's Eye View (BEV) feature map through the LiDAR encoder and BEV reduction module (flatten over $z$ dimension). For the camera stream, the images are loaded and pre-processed to a resolution of $256\times 704$. We use the SwinT~\cite{liu2021swin} backbone to process the images from $6$ cameras separately. The PV features are transformed to BEV by taking advantage of the efficient PV-to-BEV transformation module in BEVFusion~\cite{liu2023bevfusion}. The BEV feature maps from the LiDAR stream and camera stream are passed into the gated network to obtain gated feature maps with attentional relative importance between the input features. These gated feature maps are further fused by the original convolutional fusion module in BEVFusion~\cite{liu2023bevfusion}. The fused feature map is then fed into a decoder and the CenterHead~\cite{yin2021centerpoint} to generate object predictions. We train the teacher model for $20$ epochs using an AdamW optimizer~\cite{loshchilov2017adamw}. The initial learning rate is set as 2e-4 with a cosine annealing schedule~\cite{loshchilov2016cosine_annealing, mmdet3d2020}. The batch size is set as $16$. The object sampling strategy~\cite{yan2018second} is applied for the first $15$ epochs. During distillation, the pre-trained teacher weight is loaded and frozen.

\subsection{Student Model}
Similar to the LC teacher model, the gated network is also applied to the CR student model, which is denoted as BEVFusion-CR*. The stream to process the camera images is the same as the teacher model. The input radar data is processed by a PointPillar-based backbone~\cite{lang2019pointpillars, mmdet3d2020} to obtain the BEV feature map for the radar stream. The feature maps from these two streams are fused via the gated network and the convolutional fusion module in BEVFusion~\cite{liu2023bevfusion}. To maintain the consistency between the teacher and student models, the student model also uses the CenterHead~\cite{yin2021centerpoint} as the detector head. We train the student model following the same setting as the teacher model. During distillation, we load the pre-trained BEVFusion-CR* model and operate cross-modality distillation with the proposed CRKD framework.

\subsection{Base Loss Choice}
In general, $\mathcal{L}_{2}$ is more common for feature KD.
However, for CSRD, we have to consider the sensor properties. Due to radar's sparse measurements, some objects may be missed, causing radar features at corresponding locations to be outliers when computing loss with the objectness heatmap. We use $\mathcal{L}_{1}$ to downplay this effect as it penalizes large errors less heavily than $\mathcal{L}_{2}$, which leads to 0.4\% improvement in mAP than using $\mathcal{L}_{2}$. For MSFD, we follow common practice ($\mathcal{L}_{2}$) as the domain gap is relatively small (shared camera modality). For RelD, we agree with reviewer vB2P that applying $\mathcal{L}_{1}$ between similarity matrices is appropriate. For RespD, we mainly follow existing works (e.g., CMKD~\cite{hong2022cmkd}, BEVSimDet~\cite{zhao2023bevsimdet}) to choose the base loss. Our method is fairly robust to base loss choice, while the final design aligns with our design consideration and brings the best performance.

\begin{table*}[!htp]
    \centering
     \begin{tabular}{c|c|ccc|ccc}
        \toprule
        & & \multicolumn{3}{c|}{NDS$\uparrow$} & \multicolumn{3}{c}{mAP$\uparrow$}         \\
        \multirow{-2}{*}{Method} & \multirow{-2}{*}{Modality} & $[0\text{m}, 20\text{m}]$ & $[20\text{m}, 30\text{m}]$ & $[30\text{m}, 50\text{m}]$ & $[0\text{m}, 20\text{m}]$ & $[20\text{m}, 30\text{m}]$ & $[30\text{m}, 50\text{m}]$ \\ 
        \midrule
        Teacher & L+C & 76.71 & 68.63 & 50.57 & 77.11 & 62.37 & 38.25\\
        \midrule
        Student & C+R & 63.03 & 52.87 & 38.86 & 58.54 & 38.64 & 19.50\\
        \rowcolor{black!10}
        \cellcolor{white} 
        \textbf{CRKD} & C+R & \textbf{65.53(+2.50)} & \textbf{53.52(+0.65)} & \textbf{39.21(+0.35)} & \textbf{61.59(+3.05)} & \textbf{39.04(+0.40)} & \textbf{20.53(+1.03)}\\
        \bottomrule
      \end{tabular}
    \caption{Performance breakdown by range evaluated on the nuScenes \emph{val} split.
    We quantitatively show the improvement made by CRKD over the student model.}
    \label{tab:crkd_range}
\end{table*}

\section{Supplementary Experiments}

\begin{table*}[!htp]
    \centering
    \resizebox{\linewidth}{!}{
     \begin{tabular}{c|c|cccc|cccc}
        \toprule
        & & \multicolumn{4}{c|}{NDS$\uparrow$} & \multicolumn{4}{c}{mAP$\uparrow$}         \\
        \multirow{-2}{*}{Method} & \multirow{-2}{*}{Modality} & Sunny & Rainy & Day & Night & Sunny & Rainy & Day & Night\\ 
        \midrule
        Teacher & L+C & 70.22 & 71.01 & 70.54 & 44.92 & 66.02 & 65.48 & 66.25 & 41.12\\
        \midrule
        Student & C+R & 55.60  & 57.56 & 56.37 & 33.40 & 44.73 & 47.27 & 45.78 & 23.94\\
        \rowcolor{black!10}
        \cellcolor{white} 
        \textbf{CRKD} & C+R & \textbf{56.56(+0.96)} & \textbf{59.97(+2.41)} & \textbf{57.59(+1.22)} & \textbf{34.22(+0.82)} & \textbf{45.95(+1.22)} & \textbf{49.59(+2.32)} & \textbf{47.16(+1.38)} & \textbf{24.14(+0.20)}\\
        \bottomrule
      \end{tabular}
    }
    \caption{Performance breakdown by weather and lighting evaluated on the nuScenes \emph{val} split. We quantitatively show the improvement made by CRKD over the student model.}
    \label{tab:crkd_weather}
\end{table*}

\subsection{CRKD}
After loading the pre-trained weights for the teacher and student models, we add the four proposed KD loss terms to the normal detection loss and start the KD training process. We disable the object sampling strategy~\cite{yan2018second} during distillation. We set the learning rate as 1e-4 with the cosine annealing strategy and train the model for 20 epochs. The batch size is set as 8. For the loss weights, we set the hyperparameters $\lambda_1, \lambda_2, \lambda_3, \lambda_4, \lambda_5$ as $100, 10, 0.25, 1, 1$, respectively. More specifically, we set 1 as the weight for $\mathcal{L}_{respd}$ and $\mathcal{L}_{det}$ since they both compute loss for box regression and classification. The weight of $\mathcal{L}_{reld}$ is $0.25$ as it sums up the loss of $4$ downsampled affinity map pairs. We empirically select the weights of $\mathcal{L}_{csrd}$ and $\mathcal{L}_{msfd}$ (i.e., 100 and 10) to balance with other loss modules. For the Mask-Scaling Feature Distillation (MSFD), we set $r_1$ and $r_2$ as $20\text{m}$ and $30\text{m}$. The mask-scaling factors $\alpha$ and $\beta$ are set as $0.25$ and $0.5$. For the velocity threshold, we set $v_1$ and $v_2$ as $0.3\text{m/s}$ and $0.8\text{m/s}$. We also clip the object size expanding value within $[0.5\text{m}, 4\text{m}]$ to balance between different sizes of objects.

%% file: Complementary_experiments.tex
\subsection{CRKD Improvement Analysis}

Since CRKD is performing a novel KD path (LC to CR), we conduct more experiments to break down the improvement brought by CRKD to provide further insight. As the camera sensor is shared in both the teacher and student models, we narrow down our focus to the difference between LiDAR and radar integration. Radars have better long-range detection capability and weather robustness than LiDAR~\cite{RadarNet, zhou2022towards_deep_radar_survey, li2022ezfusion}. In practice, we group objects by their range to the ego-vehicle and the weather of the scene they belong to. We show mAP and NDS of the teacher model, the student model and CRKD. We highlight the quantitative improvement KD brings over the student model. As shown in~\cref{tab:crkd_range}, we can see that the most improvement comes from the short-range group. This finding demonstrates that CRKD helps the CR student detector to refine its detections in the short-range group, which can be considered as one of LiDAR's strengths as LiDAR has satisfying density for objects that are near to the LiDAR. We are also surprised to see that for mAP, the improvement in long-range group is more than the medium-range group. This finding can provide evidence that cross-modality KD can also enhance the strength of the student detector.
In addition, we group different scenes according to the weather and lighting conditions. Table~\ref{tab:crkd_weather} demonstrates increased performance from CRKD across all weather conditions, compared to the baseline student model. Notably, we see a more significant increase in improvement from CRKD in rainy weather. This finding supports that cross-modality KD can help the student to learn and leverage radar's robustness to the varying weather for better results.



\subsection{Radar Distillation Design}

\begin{table}[]
    \centering
     \begin{tabular}{c|cc|cc}
        \toprule
        Module & w/o calib & w/ calib & mAP$\uparrow$ & NDS$\uparrow$ \\
        \midrule
        ~ & \checkmark &  & 45.9 & 56.9 \\
        \cellcolor{white} 
        \multirow{-2}*{CSRD} &  & \checkmark &  46.0 & 57.0\\
        \bottomrule
      \end{tabular}
    \caption{Ablation study of CSRD with the radar calibration module.}
    \label{tab:csrd_calib}
\end{table}

\begin{table}[]
    \centering
     \begin{tabular}{c|cc|cc}
        \toprule
        Module & GT & Teacher Heatmap & mAP$\uparrow$ & NDS$\uparrow$ \\
        \midrule
        ~ & \checkmark &  & 45.9 & 56.8 \\
        \cellcolor{white} 
        \multirow{-2}*{CSRD} &  & \checkmark &  46.0 & 57.0\\
        \bottomrule
      \end{tabular}
    \caption{Ablation study of CSRD with the different distillation sources.}
    \label{tab:csrd_gt}
\end{table}

\begin{table}[]
    \centering
     \begin{tabular}{c|cc|cc}
        \toprule
        Module & max & mean & mAP$\uparrow$ & NDS$\uparrow$ \\
        \midrule
        ~ & \checkmark &  & 45.6 & 56.8 \\
        \cellcolor{white} 
        \multirow{-2}*{CSRD} &  & \checkmark &  46.0 & 57.0\\
        \bottomrule
      \end{tabular}
    \caption{Ablation study of CSRD with different channel-wise pooling methods on the heatmap predicted by the teacher model.}
    \label{tab:csrd_max_mean}
\end{table}

\begin{figure*}[]
    \centering\includegraphics[width=0.92\linewidth]{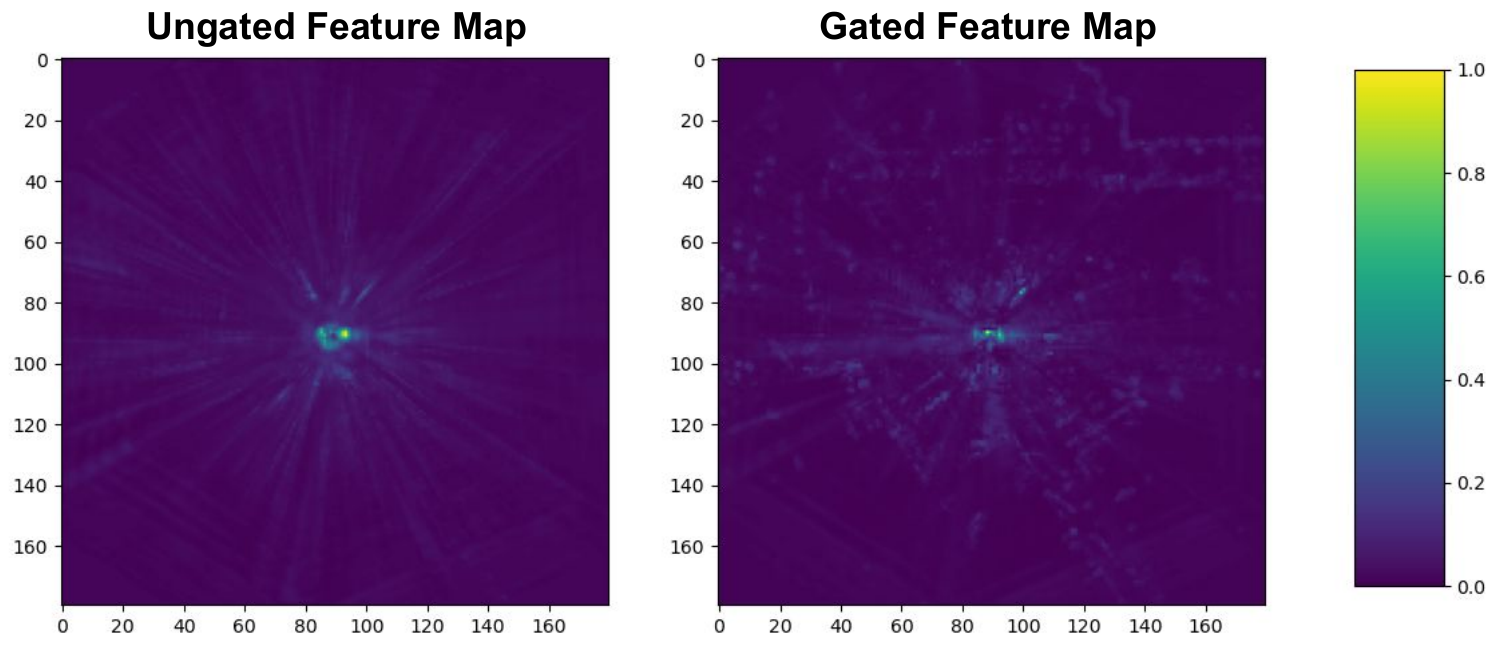}
    \caption{Visualization of the ungated and gated camera feature maps in the teacher detector. The scene geometry can be more easily interpreted from the gated feature map, as it has encoded information from the LiDAR point cloud. Best viewed in color.}
    \label{fig:gated}\vspace{-4mm}
\end{figure*}

CRKD presents a novel distillation path to a CR detector. We specifically design the KD module for radars, which has not been previously studied. We present more ablation studies to justify our design choice. We hope our work can bring more insights for future KD frameworks that leverage the radar sensor. 
In the proposed Cross-Stage Radar Distillation (CSRD) module, we design a calibration module to account for the noisy radar measurements. We conduct an ablation study to understand the effect of the calibration module. Table~\ref{tab:csrd_calib} demonstrates that the calibration module helps to further improve the performance of the student detector.

In addition to the ablation study in the main paper, we show another ablation study of the best distillation source for the CSRD module. Specifically, we compare between using the ground truth heatmap or the heatmap predicted by the teacher model. The results in~\cref{tab:csrd_gt} show that the objectness heatmap predicted by the teacher detector is a better distillation source for radar distillation.

We additionally compare taking the max or mean pooling along the class dimension of the objectness heatmap $Y^T$ predicted by the teacher detector. Table~\ref{tab:csrd_max_mean} shows that taking the mean value along different classes of the source heatmap brings more improvement.


\begin{table}[h]
    \centering
     \begin{tabular}{c|cc|cc}
        \toprule
        Module & Ungated & Gated & mAP$\uparrow$ & NDS$\uparrow$ \\
        \midrule
        ~ & \checkmark & & 45.5 & 56.8 \\ 
        \cellcolor{white}
        \multirow{-2}*{MSFD} &  & \checkmark & 45.7 & 56.9\\
        \bottomrule
      \end{tabular}
    \caption{Ablation study of MSFD with the gated camera feature.}
    \label{tab:msfd_gated}
\end{table}

\begin{table}[h]
    \centering
     \begin{tabular}{c|ccc|cc}
        \toprule
        Module & Cam & Fused & Cam\&Fused & mAP$\uparrow$ & NDS$\uparrow$ \\
        \midrule
        ~ & \checkmark & &  & 45.7 & 56.9 \\ 
        \multirow{-1}*{MSFD} &  & \checkmark & & 45.8 & 56.7\\
        \cellcolor{white}
         & & & \checkmark & 46.2 & 56.9\\
        \bottomrule
      \end{tabular}
    \caption{Ablation study of MSFD with different distillation locations.}
    \label{tab:msfd_location}
\end{table}

\begin{table}[h]
    \centering
     \begin{tabular}{c|ccc|cc}
        \toprule
        Module & Dense &  Gaussian & Ours & mAP$\uparrow$ & NDS$\uparrow$ \\
        \midrule
        ~ & \checkmark & &  & 45.7 & 56.8 \\ 
        \multirow{-1}*{MSFD} &  & \checkmark & & 45.5 & 56.7\\
        \cellcolor{white}
         & & & \checkmark & 46.0 & 57.0\\
        \bottomrule
        
      \end{tabular}
    \caption{Ablation study of MSFD with different feature masking algorithms.}
    \label{tab:msfd_mask}
\end{table}

\subsection{Feature Distillation Location}
We also experiment with introducing feature distillation at different locations. Since we introduce the gated network to the original BEVFusion~\cite{liu2023bevfusion} model, we design an ablation experiment justifying the introduction of the gated feature map to improve the feature distillation. Specifically, we compare using the gated camera feature map or the ungated camera feature map as the feature distillation source. The results shown in~\cref{tab:msfd_gated} demonstrate that the gated feature map serves as a more effective distillation source. We additionally show a qualitative example in~\cref{fig:gated} to demonstrate the benefits of using the gated feature map. The gated feature map has more informative scene-level geometry thanks to the gated network and learned relative importance weight.

Since the teacher and student models are both fusion-based, we have multiple options of feature distillation locations (e.g., camera feature, fused feature). For the proposed Mask-Scaling Feature Distillation (MSFD) module, we experiment between different locations. As shown in~\cref{tab:msfd_location}, the most effective design of MSFD is to perform the distillation of the gated camera feature map and fused feature map together. Moreover, we conduct an experiment testing alternative foreground mask generation methods. We experiment with not including any foreground mask compared to methods that include a foreground mask~\cite{liu2023bevfusion, FUTR3D, zhou2022towards_deep_radar_survey}. 
To complement the ablation study in the main paper, we compare the proposed CRKD module against the same instance without any foreground mask (denoted as dense). In addition, we try CRKD with a Gaussian-style heatmap~\cite{chen2022bevdistill, zhao2023bevsimdet}. The results are shown in~\cref{tab:msfd_mask}. This table demonstrates that though there are certain papers reporting having a Gaussian heatmap is helpful~\cite{chen2022bevdistill, zhao2023bevsimdet}, the most effective masking strategy in our scenario is still to apply the proposed mask-scaling strategy.

\subsection{Response Distillation: Strength Amplification or Weakness Mitigation?}
To better study the most suitable choice for the Response Distillation (RespD) module, we design an experiment trying to answer an insightful question: is the cross-modality distillation most helpful in amplifying the strength of the student or mitigating the weakness of the student? It is widely recognized that radars are more capable of perceiving dynamic objects~\cite{CenterFusion,li2022ezfusion, zhou2022towards_deep_radar_survey}. Therefore, the CR student may benefit from the radar's strength. As we have the flexibility of varying the loss weight $w_i$ for different classes in RespD, we experiment with different loss weight settings. In addition to the ablation study of RespD reported in the main paper, we conduct an experiment with static setting where the loss weights for static classes are set to $2$ while the weights for the other classes are set to $1$. In the static setting, priority is given to the static classes, which radars are less capable of detecting. As shown in~\cref{tab:respD}, the RespD module works better when we prioritize the learning of dynamic objects, which indicates that the RespD module is more effective when designed to be amplifying the strength of the student detector. The results complement the ablation study we show in the main manuscript, demonstrating the effectiveness of the proposed dynamic RespD module. We hope this interesting finding could provide some guidance to future study about designing cross-modality distillation to leverage the strength of different modalities effectively.

\begin{table}[]
    \centering
     \begin{tabular}{c|ccc|cc}
        \toprule
        Module & Vanilla & Static & Dynamic & mAP$\uparrow$ & NDS$\uparrow$ \\
        \midrule
        ~ & \checkmark & &  & 45.3 & 56.7 \\
        \multirow{-1}*{RespD} &  & \checkmark & & 45.4 & 56.6\\
        \cellcolor{white}
         & & & \checkmark & 45.7 & 56.7\\
        \bottomrule
      \end{tabular}
    \caption{Ablation study of Response Distillation (RespD) with different weight settings.}
    \label{tab:respD}
\end{table}

\begin{figure*}[!htp]
    \centering
    \includegraphics[width=0.9\linewidth]{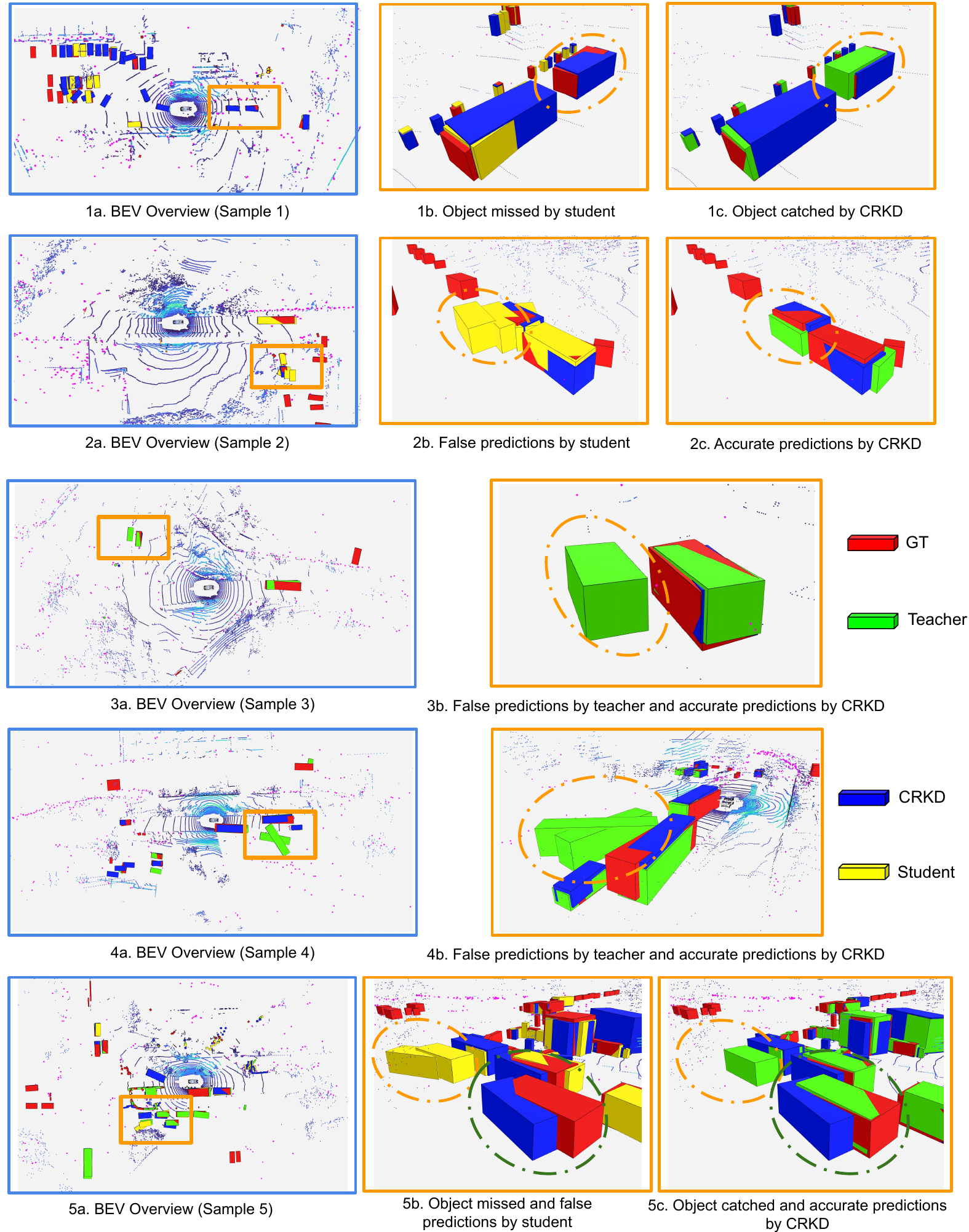}
    \caption{More Qualitative results on nuScenes. We show zoomed-in views in panel b and c for the highlighted regions in panel a, with the border dash as the correspondence. The highlighted regions are enclosed with border dash in ellipse.
    We show the ground truth annotation in {\color{red}red}, teacher prediction in {\color{green}green}, student prediction in {\color{yellow}yellow}, CRKD prediction in {\color{blue}blue}, and radar points in {\color{magenta}magenta}. In (1a) to (1c), we show an example frame where CRKD can capture the object missed by the student with the guidance of the teacher. In (2a) to (2c) we show an example frame where CRKD can reject false predictions by the student model. In (3a) to (3b) and (4a) to (4b), we show two examples where CRKD rejects false predictions by the teacher model and generates more accurate predictions. In (5a) to (5c), we show an example where CRKD outperforms both the teacher and student models by capturing missed objects and generating less false predictions.
    Best viewed on screen and in color.}
    \label{fig:supp_qualitative}
\end{figure*}

\subsection{Additional Qualitative Results}

We show additional qualitative results of CRKD in Fig.~\ref{fig:supp_qualitative}. In the first two samples (sample 1 and 2), we firstly show that CRKD is able to outperform the student model since its predictions are more aligned with the ground truth. We credit this improvement to the effective design of CRKD. We also show additional examples (sample 3 and 4) where CRKD can even outperform the teacher detector thanks to the long-range detection capability of radars. In the last sample frame (sample 5), we show that CRKD is capable of capturing the object that is missed by the student model. In addition, it is also demonstrated that CRKD is able to maintain accurate predictions where the teacher and student models generate false predictions.